\documentclass[letterpaper, 10 pt, conference]{ieeeconf}  

\IEEEoverridecommandlockouts                              

\UseRawInputEncoding

\usepackage{graphics} 
\usepackage{epsfig} 
\usepackage{times} 
\usepackage{amsmath} 
\usepackage{amssymb}  
\usepackage{tabularx,booktabs}
\usepackage[table,xcdraw]{xcolor}
\title{\LARGE \bf
Continual Learning for Class- and Domain-Incremental Semantic Segmentation
}

\author{Tobias Kalb$^{1}$, Masoud Roschani$^{2}$, Miriam 
	Ruf$^{2}$ and J\"urgen Beyerer$^{2, 3}$
\thanks{$^{1}$Tobias Kalb is with Porsche Engineering Services GmbH,
        Stuttgart, Germany
        {\tt\small Tobias.Kalb@porsche-engineering.de}}%
\thanks{$^{2}$Juergen Beyerer, Miriam Ruf and Masoud Roschani are
	with Fraunhofer IOSB, 76131 Karlsruhe, Germany {\tt\small \{masoud.roschani,
	miriam.ruf\}@iosb.fraunhofer.de}}%
\thanks{$^{3}$Juergen Beyerer is also with the Vision and Fusion Lab, Karlsruhe Institute
	of Technology KIT, c/o Technologiefabrik, Haid-und-Neu-Strasse 7,
	76131 Karlsruhe, Germany}
}

\begin{document}
\newcommand{\etal}{\textit{et al. }}

\maketitle
\thispagestyle{empty}
\pagestyle{empty}

\begin{abstract}
The field of continual deep learning is an emerging field and a lot of progress has been made. 
However, concurrently most of the approaches are only tested on the task of image classification, which is not relevant in the field of intelligent vehicles. 
Only recently approaches for class-incremental semantic segmentation were proposed. However, all of those approaches are based on some form of knowledge distillation.
At the moment there are no investigations on replay-based approaches that are commonly used for object recognition in a continual setting.
At the same time while unsupervised domain adaption for semantic segmentation gained a lot of traction, investigations regarding domain-incremental learning in an continual setting is not well-studied.
Therefore, the goal of our work is to evaluate and adapt established solutions for continual object recognition to the task of semantic segmentation and to provide baseline methods and evaluation protocols for the task of continual semantic segmentation. 
We firstly introduce evaluation protocols for the class- and domain-incremental segmentation and analyze selected approaches.
We show that the nature of the task of semantic segmentation changes which methods are most effective in mitigating forgetting compared to image classification. Especially, in class-incremental learning knowledge distillation proves to be a vital tool, whereas in domain-incremental learning replay methods are the most effective method.

\end{abstract}

\section{INTRODUCTION}

Semantic Segmentation is an essential task for environment perception in highly-automated vehicles. 
In recent years, the driving forces behind the performance gain in semantic segmentation were large-scale fully annotated datasets for semantic segmentation as well depp fully convolutional neural networks (FCN) \cite{Long_2015_CVPR}. Subsequent advances in performance for semantic segmentation were due to improvements of the FCN architecture or efficiency of the models.\\
\begin{figure}
	\includegraphics[width=0.49\textwidth]{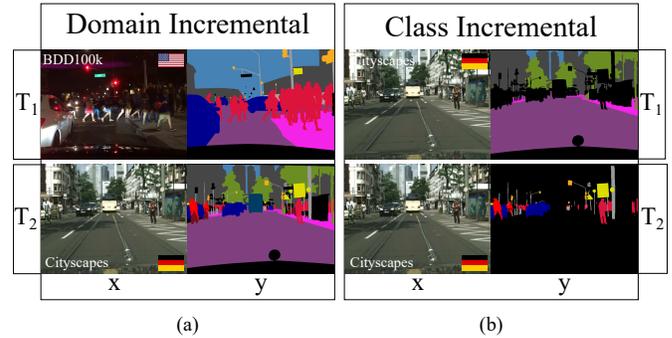}
	\caption{Domain- and class-incremental learning scenarios for semantic segmentation for a sequence of Tasks $T^{1}$ and  $T^{2}$. In the domain incremental setting only the input distribution is changing and the labels y stay the same. In the class incremental setting the input distribution stays the same, but in $T^{2}$ we observe a different set of classes.}
\end{figure}
However, a common assumption of state-of-the-art benchmarks in semantic segmentation is that all training data is collectively available. 
Additionally, it is assumed that both the number of classes is constant and that the domain of the training data matches with the target domain in which the model will operate.
However, both these assumptions severely limit the application in dynamic settings like automated driving, because (a) the driving environment is subject to constant changes and (b) it is intractable to create a training dataset that accounts for all possible driving scenarios from the beginning.
In a more realistic setting, a model starts with a very limited scope, which gradually increases over time.\\
For example, a segmentation model that has been trained for scene understanding on German urban scenes, will severely under-perform when tested on US highways, because cars, street signs, lane markings and the overall surroundings will have a different, regional specific visual appearance. 
At the same time, it is likely that we encounter new classes that we have not observed in German cities and also that some classes will not appear any more. 
For instance, it is unlikely to encounter pedestrian or bicycles on highways. 
Therefore, a model not only needs to adapt to a shift in the input domain but in the output domain as well, meaning that new classes can appear and old classes might disappear. \\
The naive solution to address this problem would be to incrementally fine-tune a model to new domains and classes using labeled data of the new domains. 
However, fine-tuning a neural network to a new domain or tasks leads to a severe performance drop of previously observed domains and tasks. 
This phenomenon is usually referred to as \emph{catastrophic forgetting} or \emph{catastrophic interference}, and is a fundamental challenge when training a neural network on a continual stream of data \cite{mccloskey1989catastrophic, Goodfellow2014}. 
In the aforementioned example it could lead to pedestrians no longer being detected by the model after it was fine-tuned to the highway scenes.\\
Significant progress has been made in addressing catastrophic forgetting in neural networks, most of the recent research has focused on the task of image classification \cite{DeLange2019, Li2018, Hsu18_EvalCL} and object detection \cite{Shmelkov2017IncrementalLO,Liu2020}, while evaluation of continual learning for segmentation is a rarely researched topic. 
Most recent approaches that tackle the challenge of continual class-incremental learning for semantic segmentation \cite{Tasar2019, Michieli2019, Klingner2020b, Cermelli2020} are all based on some form of knowledge distillation \cite{hintonKD, Li2018}. 
Currently there are no investigations on replay-based approaches that are commonly used for object recognition in a continual setting.\\
Therefore, the goal of our works is to evaluate and adapt established solutions for continual object recognition to the task of semantic segmentation and to provide baseline methods and evaluation protocols for the task of class- and domain-incremental semantic segmentation.\\


\section{Related Works}
\subsection{Continual Learning}
The commonly used categorization for continual learning of neural networks considers three different categories of methods: replay-based methods, regularization-based methods and parameter isolation methods \cite{DeLange2019}.
Replay-based methods try to retain knowledge by replaying training samples of previous learning tasks during training on new data. 
While replay might be prone to overfitting on data stored in the memory buffer and current approaches are bounded by joint training on all tasks at the same time, it has proven to be an effective method to mitigate catastrophic forgetting in continual learning \cite{Hsu18_EvalCL}.
Regularization-based methods can be divided into prior-focused and data-focused methods.
Data-focused approaches are based on the idea of knowledge distillation \cite{hintonKD} for neural networks, where the underlying goal is to transfer knowledge from a \emph{teacher} network to a \emph{student}, by propagating the outputs of the teacher as soft labels to the student. 
Learning without forgetting for example employs knowledge distillation by using the model trained on the previous task as teacher while the student is learning on new training data.
Prior-focused methods vary the plasticity of individual weights based on an estimated importance of the weight in previous tasks. A simple regularization approach for instance would be to use L2 regularization to prevent the parameters from deviating too much form parameters fitted to the previously learned tasks.
The final category of continual approaches are parameter isolation methods. 
These methods mitigate forgetting by dedicating a subset of a model's parameters to each task increment, e.g. by masking a set of parameters for each task or by growing new branches for new tasks. As these approaches usually require a task oracle at inference time, they do not satisfy the constraints of class- or domain-incremental learning \cite{DeLange2019}. 
Therefore, we do not investigate parameter isolation approaches for semantic segmentation in our experiments, for further information regarding these methods we refer to the survey by De Lange \etal \cite{DeLange2019}.

\subsection{Incremental Learning for Semantic Segmentation}

\subsubsection{Domain Adaptation}
Domain Adaption has been extensively studied for semantic segmentation especially for unlabeled data \cite{Hoffman_cycada2017, Bolte2019, vu2018advent}.
The goal in unsupervised domain adaptation is to train a model for an unlabeled target domain by transferring knowledge from a related source domain in which annotated data is accessible.
This is often achieved by learning domain-invariant features from the source and target domain, so that the classifier which is trained on the source domain can generalize to the target domain, as the features are indistinguishable.
However, these methods require that data from the source and target domain are accessible at the same time, which violates the preconditions of continual learning.
Recently, Stan et al. \cite{Stan2020} explored a more challenging setting of unsupervised domain adaption in which the source domain data is no longer available when adapting to the new domain. 
They learn a intermediate prototypical distribution from the source domain in an embedding space, which is used to minimize the discrepancy in a shared embedding space when training on the target domain.
The main difference of domain adaptation and domain-incremental learning is that domain adaptation solely aims at improving the performance on the target domain. It is not required to retain performance on the source domain. 
Currently there are no investigations on supervised domain-incremental semantic segmentation.
\subsubsection{Class-Incremental Learning} 
Most recent experiments for class-incremental object recognition have indicated that replay is one of most effective methods to mitigate forgetting \cite{Hsu18_EvalCL, 2020arXiv200909929L}, and might even be unavoidable \cite{VandeVen2020}.
However, the tasks of object recognition and semantic segmentation are of very different nature. 
One fundamental difference between the two tasks is that we observe multiple classes in a single image in semantic segmentation and, therefore, it is possible that previously learned classes reappear in new training data.

Recent approaches for class-incremental semantic segmentation use exactly this property to their advantage, by employing a teacher model that was trained on previous classes to provide soft labels for unlabeled classes in the new training data, thereby propagating previous learned representation for new training data.
For this reason, it is not required to explicitly replay samples from the previous classes as long as these classes reappear in the unlabeled parts of the new data, because the teacher provides labels for these unlabeled classes.
Tasar \etal \cite{Tasar2019} were the first to adapt a knowledge distillation loss for class-incremental semantic segmentation.
Michieli \etal and Klingner \etal \cite{Michieli2019, Klingner2020b} improved the distillation with a masked distillation loss that is only applied to the unlabeled pixels in the new training data. 
While the aforementioned methods ignore the background during training, Cermelli \etal \cite{Cermelli2020} explicitly address the issue of the semantic shift of the background class when adding new classes to the training. 
Furthermore, recently Douillard \etal \cite{Douillard2020} were the first that used hard labels in form of uncertainity-based pseudo-labels instead of the commonly used soft labels, to address the semantic shift of the background class.
These approaches significantly mitigate forgetting and do not need labels of old classes to do so. 
However, the use of knowledge distillation requires that the old classes reappear in the new training data, so that the teacher can distill knowledge of these classes during training of new classes.

Currently there are no investigations on prior-regularization approaches or replay-based methods that are commonly used for object recognition in continual setting. 
Especially replay-based methods have shown significantly better results in the class-incremental setting for object classification tasks \cite{Hsu18_EvalCL}. Also we hypothesize that replay might be required once the previously learned classes appear less frequent in new training data. 

\begin{figure*}
	\includegraphics[width=\textwidth]{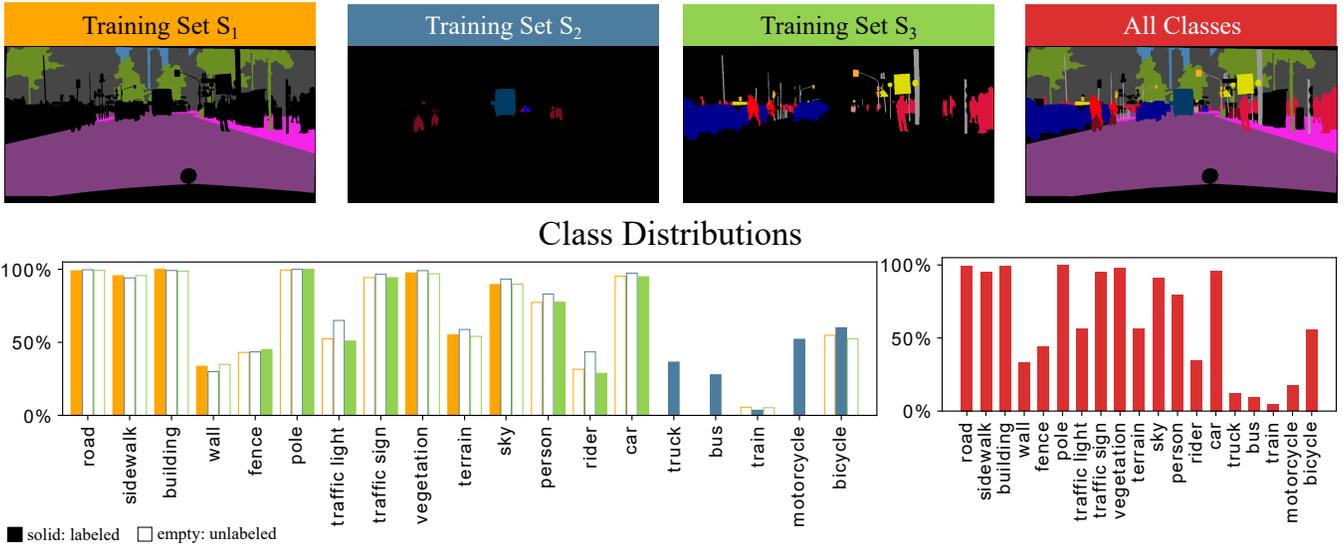}
	\caption{The three defined training subsets of Cityscapes for the class-incremental task and their respective class distributions. The solid bars in indicate that these classes are labeled in this subset while those represented in empty bars are unlabeled. The classes \textit{truck}, \textit{bus} and \textit{motorcycle} only appear in the images of training set S\textsubscript{2}. }
	\label{fig:classdistru}
\end{figure*}

\section{Incremental Semantic Segmentation}
In this section we introduce the task of semantic segmentation and categorize the different increments we can observe when training.

\subsection{Semantic Segmentation} 
A semantic segmentation task  $T$ is composed of a set $N$ pairs of input images $ x \in \mathcal{X}$ with $\mathcal{X} = \mathbb{N}^{H\times W\times 3}$ and corresponding labels $y \in \lbrace0,1\rbrace^{H\times W\times |C|}$, where $H$ and $W$ denote the spatial dimensions and $C$ is the set of semantic classes.
Given the task $T$ the goal for a model $f_{\theta}$ with parameters $\theta$, is to learn a mapping $f_{\theta} : \mathcal{X} \mapsto \mathbb{R}^{H\times W\times |C|}$ from the image space $\mathcal{X}$ to a posterior probability vector. 
For each pixel $i \in I $ with $I = \{1, \ldots,H \} \times \{1,\ldots,W\}$  in the input image $x$ and each class of $C$ the model $f_{\theta}$ estimates an posterior probability $\hat{y}_{i,c}$ that the pixel $x_{i}$ belongs to class $c \in C$.
The output segmentation masked is obtained as $\bar{y}_i = \operatorname{arg\,max}_{c\in C}\hat{y}_{i, c}$.
The model is trained by optimizing the cross-entropy between the estimated output probabilities $\hat{y}$ and the one-hot encoded ground-truth class labels $y$. It is defined by

\begin{equation}\label{eq:ce}
 \ell_{\text{ce}}(y, \hat{y}) = -\frac{1}{H\cdot W} \sum_{ i\in I }{}\sum_{c \in C}{}{y_{i, c}\log\left(  \hat{y}_{i, c} \right)} 
\end{equation}

In an incremental learning setting the model is trained on a sequence of tasks $T^k$, at incremental steps $k \in {1,2, ...}$. Therefore, we observe a new set of learning samples and output targets at each training increment $T_k$. 
The differences between two tasks $T^{k}$ and $T^{k-1}$ can be described by a shift in the label distribution $P(Y)$ and/or in the input distribution $P(X)$. 
In order to categorize the different possible task increments, we adapt the categorization proposed by Hsu \emph{et al.} \cite{Hsu18_EvalCL} and propose benchmarks for semantic segmentation for each of these settings.

\subsection{Domain-Incremental Segmentation}
In the domain-incremental setting, at each task increment the distributions of input $P(X$) changes, while the output distribution $P(Y)$ remains the same, meaning $P(X^{1}) \neq P(X^{2})$ and $P(Y^{1}) = P(Y^{2})$. In other words, we observe the same classes but they differ in appearance. 
Such a domain increment could be adapting from synthetically generated data to camera-recorded data, from one country to another, or from day to night, while the same classes would be labeled.
In order to evaluate the domain-incremental setting in our experiments, we chose to initially train the models on BDD100k \cite{bdd100k} and subsequently on Cityscapes \cite{Cordts2016Cityscapes}.

\begin{figure}
	\includegraphics[width=0.48\textwidth]{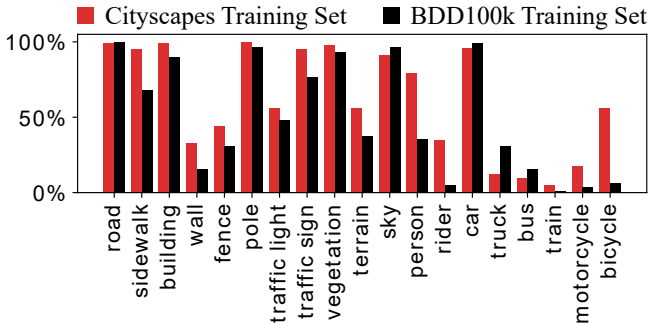}
	\caption{The class distributions $P(Y)$ of the Cityscapes and BDD100k training sets are similar and therefore are suitable for evaluation of domain-incremental training. One noticeable difference between the distributions is the higher occurrence of  the classes of rider, bicycle and motorcycle in Cityscapes. }
	\label{fig:cs_bdd_distribution}
\end{figure}

This sequence satisfies both requirements. Firstly, the requirement $P(Y^{1}) = P(Y^{2})$ is satisfied, because the datasets use the same labeling policy with a similar distributions of classes, as shown in Fig. \ref{fig:cs_bdd_distribution}. The second requirement $P(X^{1}) \neq P(X^{2})$ is satisfied as the datasets are recorded in very different driving environments. BDD100k covers very diverse driving environments including day and night drives, driving during different weather conditions, driving on highways and driving in cities. 
Cityscapes on the other hand has a much more narrow set of driving environments only covering cities during cloudy or sunny weather at daytime. 

\subsection{Class-Incremental Segmentation}
In the class-incremental setting each task introduces an exclusive subset of new classes that the model has to learn, so by definition $P(Y^{1}) \neq P(Y^{2})$.
In class-incremental \emph{image recognition} the requirement that $P(Y^{1}) \neq P(Y^{2})$ also requires $P(X^{1}) \neq P(X^{2})$, because every training sample only contains one class. However, in \emph{semantic segmentation} the input distribution can stay similar $P(X^{1}) = P(X^{2})$, as an image contains multiple classes.
For this reason, we adapt a similar approach to \cite{Klingner2020b} and divide Cityscapes into three distinct subsets. 
In each of these subsets only an exclusive selection of classes are labeled that do not reappear in the labels of the other subsets. 
However, while the classes of a set are not labeled in subsequent sets, they are likely to reappear in the images of subsequent sets, since $P(X^{1}) = P(X^{2})$.
Since this assumption is unlikely for rarely occurring objects, we increase the difficulty of the class incremental setting proposed in previous works \cite{Michieli2019,  Cermelli2020, Klingner2020b} by sampling the images in a way that three of the five classes of S\textsubscript{2} do not reappear in the training images of the remaining subsets S\textsubscript{1} and S\textsubscript{3} so that  $P(X^{1}) \neq P(X^{2})$.
The resulting class distributions of the subsets S\textsubscript{1}, S\textsubscript{2}, and S\textsubscript{3} are shown in Fig. \ref{fig:classdistru}.


\section{Experiments}

\begin{table}[]
		\caption{Comparison of DeepLabv3 (39M parameters) and ERFNet (1.2M parameters) by mean Intersection over Union (mIoU) for domain-incremental setting. Evaluation is run after training on both datasets either incremental or non-incremental.}
	\label{tab:baseline}
	\begin{tabular}{l|l|ll}
		\textbf{Method}          & Architecture & mIoU\textsubscript{BDD} & mIoU\textsubscript{CS} \\ \hline
		\textbf{Non-Incremental} & DeepLabV3    &   59.6   &    73.0  \\
		\textbf{Non-Incremental} & ERFNet       &  53.9    &    64.3  \\
		\textbf{Fine-Tuning}     & DeepLabV3    &    53.0  &    71.1  \\
		\textbf{Fine-Tuning}     & ERFNet       &   31.1   &   68.7 
	\end{tabular}
\end{table}

We hypothesize that forgetting happens faster in smaller models with less parameters. So in an effort to decrease the need to evaluate continual models on long training sequences, we evaluate the difficulty of the continual segmentation setting by training two models of different sizes.
We use the widely adapted DeepLabV3 (39M parameters) \cite{DBLP:journals/corr/ChenPSA17} and the much smaller ERFNet (1.2M parameters) \cite{Bergasa2018}.
The results in Table \ref{tab:baseline} show that both ERFNet and DeepLabV3 are capable of learning the task sequences in a non-incremental manner, and as expected DeepLabV3 achieves an absolute increase of about 8.7\% in accuracy in the non-incremental setting with respect to ERFNet. 
When learning in the incremental setting however, we observe that the drop in accuracy of ERFNet is much more severe on the previous task as compared to DeepLabV3. 
ERFNet suffers from an absolute drop of 22.8\% mIoU (mean Intersection over Union) when trained incrementally compared to the non-incremental setting, whereas DeepLabv3 only drops by an absolute of 6\% mIoU. 
This supports the initial assumption that ERFNet is far more susceptible to forgetting due to its smaller size.
Hence, we conduct all of our following experiments with the smaller ERFNet architecture, as we therefore can reduce training time significantly.

\subsection{Baseline Overview}
In our experiments we evaluate six different continual learning approaches, three of which are data-focused approaches based on knowledge distillation, two are data-focused regularization approaches and the remaining two are replay-based approaches. 

For the data-focused approaches we chose methods that are already established for semantic segmentation namely class-incremental learning (CIL) \cite{Klingner2020b} and the to semantic segmentation adapted Learning without Forgetting (LwF) \cite{Tasar2019}. LwF uses the original distillation based objective shown in EQ. \eqref{eq:kd} where the estimated probabilities $\tilde{y}$ of the teacher are used to replace the one-hot-encoded label $y$. The distillation loss is used in combination with the standard cross-entropy shown in EQ. \eqref{eq:ce}. The hyperparameter $\lambda$ is used to balance the losses.
\begin{equation}\label{eq:kd}
\ell_{\text{kd}}(\hat{y}, \tilde{y}) = -\frac{1}{|I|} \sum_{ i\in I }{}\sum_{c \in C_{k-1}}{}{\tilde{y}_{i, c}\log\left( \hat{y}_{i, c} \right)} 
\end{equation}
\begin{equation}\label{eq:lwf}
	\ell_\text{{lwf}} = \ell_{\text{ce}}(y, \hat{y}) + \lambda \ell_{\text{kd}}(\tilde{y}, \hat{y})
\end{equation}

CIL extends idea of LwF by applying the distillation loss only for unlabeled sections within the image and adding pixel-wise weights that increase the loss for classes that usually take up little space in the image.

In the second category of prior-focused regularization approaches we evaluate Memory Aware Synapses (MAS) \cite{Aljundi2018} and Elastic Weight Consolidation (EWC) \cite{Kirkpatrick2015}, both of which vary the weights' plasticity based on the estimated importance $\Omega_{i}$ of the weights in previous tasks.
The resulting loss is defined by
\begin{equation}\label{eq:reg}
	\ell_{\text{reg}}(\tilde{y}, y, \theta) =\ell_{\text{ce}}(y, \hat{y}) + \lambda \sum_{i \in \theta} \Omega_{i}(\theta_{i} - \tilde{\theta_{i}})^{2}
\end{equation}
where $\tilde{\theta_{i}}$ are the old network parameters and $\lambda$ is a hyper-parameter to balance the regularization.
EWC uses the diagonal of the Fisher information matrix to infer the importance  $\Omega_{i}$ of a weight.
MAS, on the other hand, computes the importance $\Omega_{i}$ based on how sensitive the predicted output function is to a change in this parameter $\theta_{i}$ for a new given sample. In the original MAS implementation the importance update can be also calculated on arbitrary unlabeled data, however in order to achieve a fair comparison we use the training dataset of the task at hand.
Additionally, we also use an L2 baseline which prevents parameters from deviating too far from the parameters of the previous tasks. The main difference is that all parameters are weighted equally, so that $\Omega_{i} = 1\;\forall i \in \theta $. 

The final category we evaluate in our experiments are replay-based methods. We use a naive approach which stores a randomly selected subset of images of past tasks and replays them during training on new data. We use a fixed buffer size of 32 or 64 randomly sampled images, which we store in their original resolution of $1024 \times 2048$. The memory is evenly shared for each observed task.
During training on subsequent tasks each training batch is constructed of an equal amount of new data and replay data. 

We also evaluate a combination of CIL and Replay, that employs both knowledge distillation and a small replay buffer in order to mitigate forgetting. 
As we argue that in the class-incremental setting our previously learned classes will not reappear in new training data, we need to replay these classes. We limit the replay buffer to 32 images in order to achieve a fair comparison to the naive Replay method, because for CIL+R we are also required to store the teacher model. 

Finally, in order to gauge the performance of continual learning method, the results are always compared to training the models in a non-incremental manner and by naively fine-tuning the model to new data, to infer an upper bound and lower bound performance.
When naively tuning the model to new data we evaluate two different approaches: fine-tuning and feature extraction. 
In the naive fine-tuning approach (FT) the encoder and decoder of the network are optimized, whereas for feature extraction (FE) the weights of the encoder are frozen and only the decoder head is trained.

\subsection{Experimental Setup} 
For all our experiments we use the ERFNet architecture. The encoder of the ERFNet architecture has been pre-trained on ImageNet. During class-incremental training we use a batch size of 6, whereas for the domain-incremental setting we use a batch size of 16. As proposed by Bergasa \etal \cite{Bergasa2018} we train the model using the Adam optimizer with a momentum of 0.9, weight decay of $3 \times 10^{-4}$. 
Moreover, we use a learning rate of $4\times 10^{-4}$ and a polynomial learning rate schedule with a power of 0.9.
We use early stopping on a hold-out set of the current tasks and train for a maximum of 250 epochs in all our experiments. 
During training we crop the images to a 2:1 ratio, scale the images with a factor between 1.0 and 2.0 and finally crop the image to a random region of size $512 \times 1024$. 
During testing we use the original image sizes.
We keep the network architecture unchanged except for regularization approaches and fine-tuning, where we lower the learning rate to $1\times 10^{-5}$ and freeze the Batch Normalization layers of the network after training on the first task. Additionally, in the class incremental setting we add a new decoder head for the new classes, as proposed by \cite{Klingner2020b}.

All accuracy results are reported using the mean Intersection over Union (mIoU) on the respective evaluation sets of Cityscapes and BDD. In order to evaluate approaches in the class-incremental setting we compute the mIoU only in regard to the classes of each subset, so when evaluating on $S_{1}$ we do not account for classes of $S_{2}$ or $S_{3}$ . 
However, after completing training on the final subset $S_{3}$, we compute the mIoU\textsubscript{Cityscapes} for all 19 classes on the Cityscapes evaluation set.

\renewcommand{\arraystretch}{1.15}
\begin{table*}[]
	\centering
	\caption{Comparison of Continual Methods on ERFNet by mean Intersection over Union (mIoU) for the class-incremental setting. Evaluation is run on the Cityscapes validation set using the indicated class subsets shown in Fig. \ref{fig:classdistru}.}
	\label{tab:ci_res}
	\begin{tabularx}{0.75\textwidth}{l|cc|cccc}
		\textbf{}       & \multicolumn{2}{c|}{Evaluation after Training on $S_1$, $S_2$} & \multicolumn{4}{c}{Evaluation after Training on $S_1$, $S_2$, $S_3$} \\ \hline
		\textbf{Method} & $\text{mIoU}_{S_1}$  & $\text{mIoU}_{S_2}$     & $\text{mIoU}_{S_1}$               & $\text{mIoU}_{S_2}$                & $\text{mIoU}_{S_3}$              & $\text{mIoU}_{\text{Cityscapes}}$               \\ \hline
		Non-Incremental         	      & $-$               & $-$                  & 83.8            & 76.2             & 81.0            & 68.7         \\\hline
		FT    				      & 32.6             & 58.1               & 32.8            & 12.0             & 79.9			 & 16.1         \\ 
		FE 					      & \textbf{79.2}    & 49.0               & \textbf{79.2}   & 49.0             & 64.8            & 39.6         \\\hline
		L2              	      & \underline{79.2} & 49.5               & \underline{78.7}& 48.4             & 65.0            & 39.8             \\
		MAS \cite{Aljundi2018}    & 49.4             & 60.0               & 53.2            & 40.4             & 76.6            & 30.5             \\
		EWC \cite{Kirkpatrick2015}& 26.3             & 58.9               & 46.8            & 33.6             & 79.3            & 25.2             \\ \hline
		CIL \cite{Klingner2020b}  & 78.3             & \textbf{69.8}      & 78.1            & 42.3             & \underline{82.0}& \textbf{56.6}         \\
		LwF \cite{Tasar2019}      & 77.2             & \underline{68.3}   & 76.4            & 38.1             & \textbf{82.6}   & 53.9         \\ \hline
		CIL+R $n=32$              & 78.3             & 64.6      		  & 77.4            & \textbf{43.3}    & \underline{82.0}& \underline{56.0} \\   
		Replay $n=32$             & 67.7             & 59.2               & 67.3            & 38.0             & 80.8            & 39.8         \\  
		Replay $n=64$             & 69.3             & 58.0               & 68.4            & \underline{42.6} & 79.8            & 39.8         \\

	\end{tabularx}
\end{table*}


%




\section{Results and Discussion}
The results of the class-incremental and domain-incremental experiments are shown in Table \ref{tab:ci_res} and \ref{tab:di_res}. 

\subsection{Results on class-incremental learning}
First of all, we look at the results of the baseline approaches FT and FE to indicate what we can expect in terms of adaptability when training on new classes. 
The results in Table \ref{tab:ci_res} show that FT is better at adapting to new classes than FE, but suffers from severe forgetting on old classes, dropping from $\text{mIoU}_{S_2} = 58.1\%$ to only $\text{mIoU}_{S_2}= 12.0\%$ after training on $S_3$.
On the other hand, the feature extraction (FE) approach is able to retain the performance on previously trained classes, but is not able to adapt to new classes, due to the fixed weights in the encoder.
The L2 regularization approach performs similar to the FE approach in terms of mitigating forgetting of previous classes. However, since the weights in the encoder are no longer completely fixed, the model is better at adapting to new classes, resulting in small increase for the latest classes. 
The remaining regularization approaches MAS and EWC perform similar to the basic FE or FT approaches. Furthermore, the L2 baseline achieves an overall higher accuracy than EWC and MAS.
Hsu \etal \cite{Hsu18_EvalCL} observed a similar result in the evaluation for class-incremental image recognition tasks.
It is also striking that all of the aforementioned approaches score a high mIoU on the individual subsets, but achieve an unexpectedly low mIoU\textsubscript{Cityscapes} when evaluated on all classes at the same time.
This indicates that in these approaches the model does not learn to effectively distinguish between classes of the individual subsets.
Only when replaying classes from the past sequences, either by knowledge distillation (CIL, LwF) or replaying samples from the past (naive Replay), the model learns to distinguish between those classes on the final evaluation set.
CIL and LwF both achieve similar results as the replay methods when they are being evaluated on the individual subsets, without the need to store additional data.
However, we observe a significant boost in a accuracy for CIL and LwF when we evaluated on all classes at the same time, resulting of an absolute increase of 16.8\% mIoU\textsubscript{Cityscapes} compared to naive Replay.
Finally, the drop in accuracy of the $\text{mIoU}_{S_2}$ after training on $S_3$ show that distillation-based methods suffer from forgetting for classes that do not reappear in subsequent datasets as the knowledge for these classes cannot be replayed. Combining CIL and Replay shows helps to mitigate the effect of forgetting the classes of $S_2$, the effect however is minor due to the small buffer size.

The overall results show that contrary to the results for \emph{image classification} that were reported by Hsu \etal \cite{Hsu18_EvalCL} knowledge distillation is the most effective method for class incremental segmentation. However, some form of replay will be required for classes that will not reappear in new training data to prohibit forgetting of these classes.

\subsection{Results on domain-incremental learning}
The results for domain-incremental setting are illustrated in Table \ref{tab:di_res}. 
First of all, the results show that all of the approaches mitigate forgetting in some way, but particularly the regularization-based approaches struggle to adapt to new data, which leads to a decrease on the overall accuracy mIoU\textsubscript{average}.
The prior-focused regularization approaches barely outperform simple fine-tuning of the model. 
Similar to the results in the class-incremental setting, EWC is the best prior-focused regularization approach at adapting to the new data, while the L2 regularization is best at retaining performance of the previous task. 
LwF outperforms the prior-focused methods (EWC, MAS, L2) on previous task performance but at the same time drops 7.8\% mIoU\textsubscript{Cityscapes} on the new task compared to simple fine-tuning.
This leads to the conclusion that the teacher-model limits the student's ability to adapt to new data. 
This happens when the teacher's predictions contradict the current training labels. 
In the class-incremental setting, this problem does not arise, since the teacher's predictions are for a disjoint set of classes that occur in a different region of the image, which is why knowledge distillation is very effective in the class-incremental setting.
Naive Replay achieves the highest accuracy on both datasets reaching an mIoU\textsubscript{average}=56.1\%, which is 3\% absolute below the upper-bound performance of the offline model. 
It should be noted that in this case the replay method is only using 32 images of the original 7000 (0.46\%) images of BDD100k during training on Cityscapes. But since we compose each batch using data from memory, we condition the model to retain the domain-specific features it has previously learned.
Finally, the combination of using knowledge distillation and naive replay does not provide any benefits compared to using replay on its own.
\begin{table}[]
	\centering
	\caption{Comparison of Continual Methods on ERFNet by mean Intersection over Union (mIoU) for the domain-incremental setting. Evaluation is run after training on both datasets either incremental or non-incremental.}
	\label{tab:di_res}
	
	\begin{tabular}{l|ccc}
		
		\textbf{}       & \multicolumn{3}{c}{Evaluation after Training on BDD, Cityscapes}\\ \hline
		\textbf{Method} & \textbf{mIoU\textsubscript{BDD}} & \textbf{mIoU\textsubscript{Cityscapes}} & \textbf{mIoU\textsubscript{average}}                  \\ \hline
		Non-Incremental        		& 53.9                 & 64.3             &  59.1\\
		FE               	   		& 36.9                 & 54.3             &  45.6                          \\ 
		FT                     		& 31.1                 & 68.7             &  49.9                          \\ \hline
		L2                   		& 37.4                 & 55.6             &  46.6                              \\
		MAS \cite{Aljundi2018}      & 34.3                 & 54.8             &  44.5                             \\
		EWC \cite{Kirkpatrick2015}  & 34.4                 & 67.0             &  50.7                             \\\hline
		LwF \cite{Tasar2019}        & 39.3                 & 60.9             &  44.0                             \\\hline
		Replay $n=32$     			& 42.8                 & \textbf{69.2}    &	\underline{56.0}\\ 
		Replay $n=64$    			& \textbf{44.6}        & \underline{67.6} &	\textbf{56.1}						 \\ 
		LWF+R $n=32$      			& \underline{43.1}     & 62.1             &  52.6                                                      
	\end{tabular}
\end{table}
\subsection{Comparison to Results on Image Recognition}
The results show, that not all insights that apply to continual image classification also hold for semantic segmentation. We will now compare our results with the findings of Hsu \etal \cite{Hsu18_EvalCL}, who thoroughly evaluated a broad range of continual learning approaches for image recognition.
The most obvious similarities, are firstly the failure of regularization approaches to effectively mitigate forgetting and secondly the effectiveness of naive Replay in the domain-incremental setting. However, in the class-incremental setting due to the nature of semantic segmentation that multiple classes appear within one image, the effectiveness of other approaches shifts. Most notably knowledge distillation is far more effective as classes are likely to reappear in new training images achieving the best results in our evaluation and barely outperform fine-tuning in image recognition \cite{Hsu18_EvalCL}. Furthermore, in our evaluation replay-based methods have shown to struggle when distinguishing between classes of different class-increments, while for image recognition replay is often seen as a requirement in the class-incremental setting \cite{Hsu18_EvalCL, VandeVen2020}.


\section{Conclusion}
We compare and adapt established methods for continual learning to the task of semantic segmentation and investigate their effectiveness in a class- and domain-incremental learning. 
The results show that the nature of the task of semantic segmentation changes which methods mitigate forgetting in continual learning compared to image classification. 
In particular, knowledge distillation has proven to be a vital tool to mitigate forgetting in the class-incremental setting. 
Furthermore, none of the analyzed methods provided good results for class- and domain-incremental learning. While knowledge distillation excels in the class-incremental setting, it is not suited for domain-incremental learning as it struggles to adapt to new data. On the other hand, naive Replay is the most effective method to overcome forgetting in domain-incremental learning only using 32 images (0.46\%) of the previously observed training data, but struggles to distinguish between old and new classes in the class-incremental setting. 
Future directions will investigate how the combined class- and domain-incremental learning setting can be a addressed, in which the appearance of old classes can change while we observe new classes at the same time.
%

\bibliographystyle{IEEEtran}
\bibliography{IEEEabrv,mybibfile}

\addtolength{\textheight}{-12cm}   




\section*{ACKNOWLEDGMENT}

The research leading to these results is funded by the Federal Ministry for Economic Affairs and Energy of Germany within the project ``KI Delta Learning -- Development of methods and tools for the efficient expansion and transformation of existing AI modules of autonomous vehicles to new domains''. The authors would like to thank the consortium for the successful cooperation.

\end{document}